\documentclass[a4paper, journal]{IEEEtran}
\IEEEoverridecommandlockouts
\usepackage{cite}
\usepackage{amsmath,amssymb,amsfonts}
\usepackage{algorithmic}
\usepackage{graphicx}
\usepackage{xcolor}
\usepackage{tabularx,booktabs,threeparttable}
\newcolumntype{C}{>{\centering\arraybackslash}X} 
\usepackage{lipsum}
\usepackage{graphicx}
\usepackage{textcomp}
\usepackage{multirow}
\def\BibTeX{{\rm B\kern-.05em{\sc i\kern-.025em b}\kern-.08em
    T\kern-.1667em\lower.7ex\hbox{E}\kern-.125emX}}
\usepackage{fancyhdr}

\begin{document}

\title{Land Cover and Land Use Detection using Semi-Supervised Learning}
\author{ 
    \IEEEauthorblockN{Fahmida Tasnim Lisa\textsuperscript{*}, Md. Zarif Hossain\textsuperscript{*}, Sharmin Naj Mou\textsuperscript{*}, Shahriar Ivan, and Md. Hasanul Kabir\\}

    \IEEEauthorblockA{Department of Computer Science and Engineering,\\Islamic University of Technology, Gazipur, Bangladesh\\}
    
    \IEEEauthorblockA{Email: \{fahmidatasnim, zarifhosssain, sharminnaj, shahriarivan, hasanul\}@iut-dhaka.edu}
}

\maketitle
\fancypagestyle{firstpage}
{
    \fancyhead[L]{\small\small{2022 25th International Conference on Computer and Information Technology (ICCIT) \\ 17-19 December, Cox’s Bazar, Bangladesh}}    
    \fancyfoot[L]{ \small{979-8-3503-4602-2/22/\$31.00 \copyright 2022 IEEE}}
    \fancyfoot[R]{\small{* These authors contributed equally}}
    \cfoot{}
}
\thispagestyle{firstpage}
\begin{abstract}
Semi-supervised learning (SSL) has made significant strides in the field of remote sensing. Finding a large number of labeled datasets for SSL methods is uncommon, and manually labeling datasets is expensive and time-consuming. Furthermore, accurately identifying remote sensing satellite images is more complicated than it is for conventional images. Class-imbalanced datasets are another prevalent phenomenon, and models trained on these become biased towards the majority classes. This becomes a critical issue with an SSL model's subpar performance. We aim to address the issue of labeling unlabeled data and also solve the model bias problem due to imbalanced datasets while achieving better accuracy. To accomplish this, we create ``artificial" labels and train a model to have reasonable accuracy. We iteratively redistribute the classes through resampling using a distribution alignment technique. We use a variety of class-imbalanced satellite image datasets: EuroSAT, UCM, and WHU-RS19. On UCM balanced dataset, our method outperforms previous methods MSMatch and FixMatch by 1.21\% and 0.6\%, respectively. For imbalanced EuroSAT, our method outperforms MSMatch and FixMatch by 1.08\% and 1\%, respectively. Our approach significantly lessens the requirement for labeled data, consistently outperforms alternative approaches, and resolves the issue of model bias caused by class imbalance in datasets.
\end{abstract}

\section{Introduction}
There have been an abundant satellite and aerial remote sensing images since significant advancements in earth observational technology have been made. Plenty of satellites are orbiting the globe, collecting massive amounts of remote sensing data. Furthermore, the quantity is increasing at an exponential rate every day. According to reports, 1,029 small satellites (under 600 kg) were launched in 2020 alone, setting a record \cite{erwin_2020}. Since the quantity of remote sensing data is growing yearly, so is the demand for effective image processing methods. For training neural networks, deep learning approaches rely substantially on labeled data. Typically, experts manually annotate these labeled data, which is scarce\cite{sydorenko_2020} because manually annotating data is time-consuming and expensive. Typically, a remote-sensing image contains a variety of distinct ground objects. So, it is difficult for the model to identify a scene from a remote sensing image that contains a variety of ground items to achieve a suitable level of accuracy \cite{cheng2020remote}. There are also many more pixels and resolutions in satellite pictures than in conventional pictures. 
Semi-supervised land scene classification can correctly generate labels for land cover and land use detection on remote sensing data with a limited number of labels and reach a reasonable level of accuracy. Semi-supervised learning (SSL) uses a small amount of labeled data and a huge amount of unlabeled data to train a neural network. There have been significant advances in semi-supervised learning (SSL) in recent years \cite{sohn2020fixmatch,berthelot2019mixmatch,berthelot2019remixmatch}. These methods reduce time and save costs while simultaneously achieving the same level of accuracy as fully supervised methods in similar scenarios.\

Moreover, there have been few works where semi-supervised learning has been applied to process satellite images \cite{liu2017semi} \cite{wu2017semi}. However, the satellite image datasets that are available\cite{helber2019eurosat,helber2018introducing,xia2017aid,yang2010bag,dai2010satellite,xia2010structural,long2021creating} are class-imbalanced, with the majority classes having a large number of samples and the minority classes having a smaller sample size.
Models trained on data with an uneven distribution of classes become biased towards the majority classes, which is a major cause of an SSL model's poor performance. There have been studies to solve model bias, such as re-weighting and averaging \cite{buda2018systematic}. These techniques, however, require labeled samples and are thus dependent on them. SSL's impact on imbalanced data has not been thoroughly investigated, making it a newcomer to the scene.  \cite{wang2019semi,mayer2021adversarial,kim2020distribution,chen2020semi,he2021re,wei2021crest} have all worked on handling imbalanced data utilizing distribution alignment with SSL algorithms.\

In this paper, we introduce a semi-supervised learning approach based on \cite{sohn2020fixmatch} and a distribution alignment strategy\cite{wei2021crest}, to address the problem of labeling land use and land cover images and to address the problem of class imbalance. We compare our results with two semi-supervised learning techniques, MSMatch \cite{gomez2021msmatch} and FixMatch\cite{sohn2020fixmatch} (with tweaked augmentation) on three datasets, EuroSAT \cite{helber2019eurosat} \cite{helber2018introducing}, UC Merced
Land Use (UCM) dataset \cite{yang2010bag} and WHU-RS19 \cite{dai2010satellite} \cite{xia2010structural}. Our proposed technique combines a class rebalancing technique to retrain an enhanced semi-supervised learning model with a unique augmentation strategy that incorporates collecting pseudo-labeled data from the unlabeled set to increase the samples of the minority classes of the original labeled set. Each completely trained SSL model is called a \textit{generation}.

The pseudo-labeled samples are extracted from the unlabeled set and put into the labeled set after each \textit{generation} to retrain the semi-supervised learning model with tweaked augmentation. We employ a stochastic update technique instead of updating the labeled set with all pseudo-labeled samples generated from the SSL model. The samples are picked with a high probability (threshold crossing 95\%) only if they are from minority classes since they are more likely to be accurate predictions. The updating probability is determined by the data distribution produced from the labeled set. In this way, our proposed method decreases pseudo-labeling bias and improves the test set's accuracy. We show that our suggested technique outperforms FixMatch \cite{sohn2020fixmatch} under the custom augmentation by 2.32\% accuracy in the case of the UCM dataset, 1\% improvement in the case of EuroSAT, and 2.31\% in the case of WHU-RS19 dataset.

The remainder of this article is organized as follows: Recent advancements in Semi-Supervised Learning, different approaches to Semi-supervised learning, and Distribution Alignment used in Remote Sensing Scene Classification are discussed in Section \ref{sec:rel_works}. Section \ref{sec:method} provides an overview of our proposed method, FixMatch, and Augmentation. Section \ref{sec:result} presents the datasets used in our experiment, a comparison of our proposed method with other methods, and the training. Section \ref{sec:conclusion} presents our research's conclusion and discusses our work plan.This paper introduces a semi-supervised learning approach based on \cite{sohn2020fixmatch} and a distribution alignment strategy\cite{wei2021crest} to address the problem of labeling land use and land cover images and to address the problem of class imbalance. We compare our results with two semi-supervised learning techniques, MSMatch \cite{gomez2021msmatch} and FixMatch\cite{sohn2020fixmatch} (with tweaked augmentation) on three datasets, EuroSAT \cite{helber2019eurosat} \cite{helber2018introducing}, UC Merced
Land Use (UCM) dataset \cite{yang2010bag} and WHU-RS19 \cite{dai2010satellite} \cite{xia2010structural}. Our proposed technique combines a class rebalancing technique to retrain an enhanced semi-supervised learning model with a unique augmentation strategy that incorporates collecting pseudo-
labeled data from the unlabeled set to increase the samples of the minority classes of the original labeled set. Each completely trained SSL model is called a \textit{generation}.

The pseudo-labeled samples are extracted from the unlabeled set and put into the labeled set after each \textit{generation} to retrain the semi-supervised learning model with tweaked augmentation. Instead of updating the labeled set with all pseudo-labeled samples generated from the SSL model, we employ a stochastic update technique. The samples are picked with a high probability (threshold crossing 95\%) only if they are from minority classes since they are more likely to be accurate predictions. The updating probability is determined by the data distribution produced from the labeled set. In this way, our proposed method decreases pseudo-labeling bias and improves the test set's accuracy. We show that our suggested technique outperforms FixMatch \cite{sohn2020fixmatch} under the custom augmentation by 2.32\% accuracy in the case of the UCM dataset, 1\% improvement in the case of EuroSAT, and 2.31\% in the case of WHU-RS19 dataset.

The remainder of this article is organized as follows: Recent advancements in Semi-Supervised Learning, different approaches to Semi-supervised learning, and Distribution Alignment used in Remote Sensing Scene Classification are discussed in Section \ref{sec:rel_works}. Section \ref{sec:method} provides an overview of our proposed method, FixMatch, and Augmentation. Section \ref{sec:result} presents the datasets used in our experiment, a comparison of our proposed method with other methods, and the training. Section \ref{sec:conclusion} presents the conclusion of our research and discusses our future plan of work.
\vspace{-3mm}
\section{Related Works}\label{sec:rel_works}

\subsection{Recent Advancements in Semi-supervised Learning}

A serious limitation in deep learning methods is the need for labeled data when training a neural network.
Pseudo-labeling \cite{lee2013pseudo} is a semi-supervised learning method that involves creating ``artificial" labels to label the unlabeled data. 
In a dataset with a lot of unlabeled data, \cite{jeong2019consistency} used a consistency regularization (CR) based method to improve image object detection.
MixMatch \cite{berthelot2019mixmatch} is a semi-supervised learning approach that combines traditional regularization, consistency regularization, and entropy minimization.  
To increase the amount of data, the unlabeled and labeled data are combined and mixed using MixUp\cite{zhang2017mixup} before being fed to the model to increase accuracy.
FixMatch \cite{sohn2020fixmatch} is an algorithm that combines consistency regularization and pseudo-labeling. FixMatch achieves cutting-edge performance across a range of benchmarks despite its simplicity. With just 250 labels for 10 classes, it achieves 94.3\% accuracy on the CIFAR-10\cite{krizhevsky2009learning}, and with only 40 labels (4 data per class), it achieves 88.61\% accuracy. ReMixMatch \cite{berthelot2019remixmatch} is an improved version of MixMatch \cite{berthelot2019mixmatch}.

\subsection{Semi-supervised Learning for Remote Sensing Scene Classification}

There is a plethora of data on remote sensing images that are freely available. A variety of machine learning and deep learning algorithms have been utilized to make use of these data \cite{jia2021survey}\cite{ma2019deep}. Convolutional neural network (CNN) based remote sensing image classification \cite{liu2017semi}, auto-encoder based remote sensing image classification, and generative adversarial network (GAN) based remote sensing image classification have all been used in the past. For remote sensing image classification, supervised, self-learning and semi-supervised learning have all been used. \cite{othman2016using} proposed a remote sensing image scene classification approach based on convolutional features and a sparse autoencoder in their work. Even though autoencoder-based remote sensing image classification has achieved good results, these techniques are unable to fully exploit scene class information.
Many GAN-based remote sensing image classifications in a semi-supervised manner have been done \cite{teng2019classifier}  \cite{ma2019siftinggan}.
A new strategy for learning discriminative convolutional neural networks (D-CNNs) was proposed in an earlier work \cite{cheng2018deep}. \cite{cheng2020remote} investigated remote sensing image classification using auto-encoders, GANs, and CNNs. For better performance and results, these CNN-based algorithms require annotated images, which is where SSL comes in. \cite{wu2017semi} proposed using deep convolutional recurrent neural networks to employ deep learning for hyperspectral image classification. \cite{cenggoro2017classification} used variational SSL to classify imbalanced satellite data.
\cite{fan2020semi} used Submeter HRRS Images. They proposed a semi-supervised learning strategy in addition to multiple deep learning-based CNNs.
\subsection{Distribution Alignment}
The distribution that the model predicted in the case of unlabeled data is aligned with the class distribution of the labeled training set by the distribution alignment (DA) method \cite{berthelot2019remixmatch}, which is especially well adapted to class-imbalanced situations. The feature distribution alignment approach \cite{mayer2021adversarial} performs exceptionally well in overcoming SSL overfitting due to distribution mismatch in samples. The technique of augmented distribution alignment \cite{wang2019semi} used an adversarial training approach to decrease the distribution distance of labeled and unlabeled data and generated pseudo training samples to solve the labeled data's small sample size issue. Based on the information collected through clustering, the FeatMatch \cite{kuo2020featmatch} feature-based refinement and augmentation algorithm prevented the need for significant additional processing by storing features computed across iterations in memory. The semantic pseudo-label and the linear one class are compatible with the DASO method \cite{oh2021distribution}. Uncertainty-Aware Self-Distillation (UASD)\cite{chen2020semi}, a different method, creates soft targets that stop catastrophic error propagation and allow learning from unconstrained unlabeled data, including out-of-distribution (OOD) samples.

DARS \cite{he2021re} produces unbiased pseudo-labels where the pseudo-labels allow for a match with the true class distribution of the labeled data. In DARP \cite{kim2020distribution}, a convex optimization problem is formulated, and a simple iterative algorithm is developed to correct the pseudo-labels that are generated from a biased model. A baseline SSL model is repeatedly retrained using CReST \cite{wei2021crest}, where it makes use of a labeled set that is increased by adding samples from an unlabeled set that has been pseudo-labeled. Following an estimated class distribution, minority class pseudo-labeled samples are chosen more frequently in this study.
 \section{Method}\label{sec:method}

  \subsection{Overview}
Our proposed method aims to provide good accuracy on the imbalanced satellite datasets, and it works to eliminate model bias to the majority classes while rebalancing the dataset with each \textit{generation}.

\begin{figure}[h]
    \centering
    \includegraphics[width=0.45\textwidth]{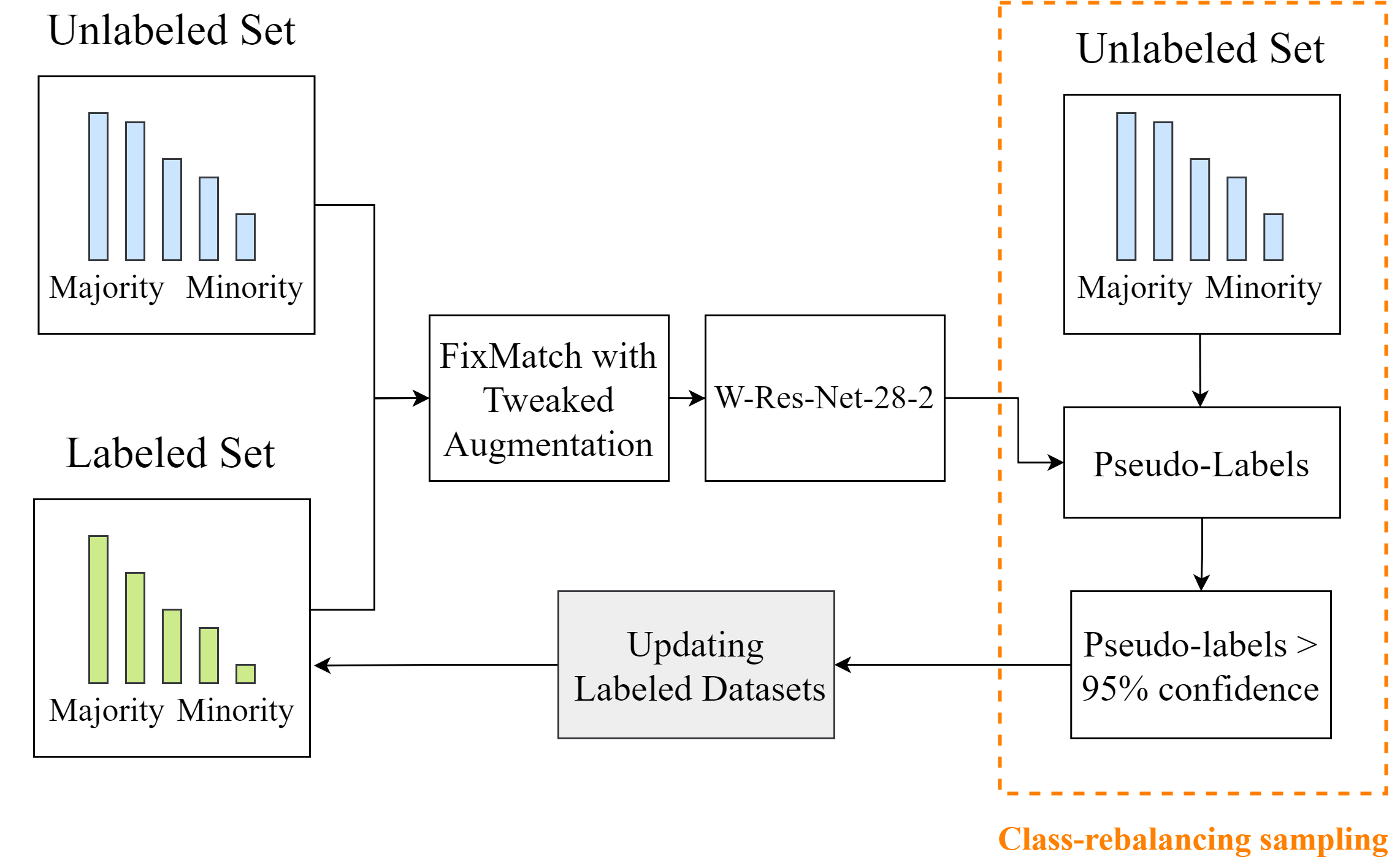}
    \caption{Proposed Method: (a) Model trained using baseline SSL algorithm, Fixmatch \cite{sohn2020fixmatch} (b) Pseudo labeling is done on unlabeled data and class-rebalancing is applied with multiple \textit{generations}}
    \label{fig:img1}
\end{figure}

Figure \ref{fig:img1} shows a basic overview of our pipeline. A base SSL algorithm is initially applied, and the model is trained using both the available labeled and unlabeled data. Here, Fixmatch and our modified augmentation are used as the basic SSL algorithm. The photos from the unlabeled dataset are then pseudo-labeled in the following phase after the model has proper confidence. We proceed to the last stage if the pseudo labels confidence level meets the predefined threshold. To rebalance our dataset, in this step, high precision pseudo-labels from minority classes are chosen with a higher sampling rate, while a lower sampling rate is used for pseudo-labels from majority classes. This completes one \textit{generation}.
\vspace{-2mm}
 \subsection{FixMatch}
Fixmatch \cite{sohn2020fixmatch} is an advanced semi-supervised learning (SSL) algorithm that combines consistency regularization and pseudo-labeling to achieve greater performance on a few available data. Pseudo-label is the process of automatically labeling unlabeled data using a learned model. Moreover, consistency regularization states that the model should predict the same output for similar inputs. 

The model is initially trained using the available labeled data. Then the process is executed in two pipelines, where first, a weakly-augmented image is inserted into the model. When the model assigns a probability to any class and exceeds the threshold, the predictions from weakly augmenting labeled data are converted into a one-hot pseudo-label. Then, we used a cross-entropy loss with pseudo-labels to train the model to make predictions on the strongly-augmented version of unlabeled images. In this case, we treat the pseudo-label as the true label and try to minimize the loss.

We utilized the two types of augmentation strategies of FixMatch: ``weak" and ``strong."
Standard flip-and-shift and random crop were used for the weak augmentation, while RandAugment \cite{cubuk2019randaugment} was used for the strong augmentation strategy.
RandAugment utilizes Auto Contrast, Brightness, Contrast, Invert, Rotate, Color, Hue, and many other transformation functions. To improve the results on remote sensing datasets, we adjusted the augmentation parameters. This enabled us to improve over the baseline FixMatch algorithm and enhance the accuracy of satellite imagery.
 \vspace{-2mm}
 \subsection{Augmentation}
 Augmentation is widely used to help neural networks generalize better to unseen images \cite{cubuk2019randaugment}. The consistency Regularization strategy of FixMatch  \cite{sohn2020fixmatch} gets leveraged with the help of two different kinds of augmentation.

The satellite photos were not a suitable fit for FixMatch's original augmentation parameters, so the results were relatively poor. This is because the satellite images were losing too much information due to the heavy augmentations of FixMatch, and since most satellite images included fog and other visibility issues, they were already fairly hazy. Therefore, we modified FixMatch's augmentation process and parameters to resolve this issue.
 
 \begin{table}[h]
\centering
\caption{Tweaked Augmentation strategy for Satellite Images}
\resizebox{7cm}{!}{
\begin{tabular}{|l|l|l|}
\hline
\textbf{Transformation} & \textbf{Tweaked \tnote{a}} &  \textbf{Parameter Range} \\ \hline
Auto Contrast            &                          &                       \\ \hline
Brightness                         &     \checkmark         & [0.1, 0.2]    \\\hline
Color                        &                    &    [0.05, 0.95] \\ \hline
Hue                        &                    &    0.1 \\ \hline
Equalize                        &                           &                         \\ \hline
Identity                        &                          &                         \\ \hline
Posterize                      &   \checkmark                          &      \\ \hline
Shift                      &                         &   [0.1,0.2]   \\ \hline

Rotate                        &                           &   [-30, 30]                      \\ \hline
Sharpness                       &     \checkmark                      &    [0.5, 1]
                     \\ \hline
Shear x                        &     \checkmark                      &    [0.1, 0.2]
                     \\ \hline
Shear y                        &      \checkmark                     &     [0.1, 0.2]
                    \\ \hline
Solarize                        &       \checkmark                     &                       \\ \hline

Translate x                        &             & [0, 1]                                     \\ \hline
Translate y                        &           & [0, 1]                                    \\ \hline
\end{tabular}
}
\smallskip
\scriptsize
\begin{tablenotes}
\item[a] \checkmark: Augmentations that have been tweaked
\end{tablenotes}

\label{table:aug_fix}
\end{table}
 
Here, Table \ref{table:aug_fix} shows all the augmentations we used in our model along with the changes in the augmentation parameters of the strong augmentation strategy, i.e., the RandAugment\cite{cubuk2019randaugment} to make it suitable for the data we are working with. The augmentations we have tweaked are shown with checkmarks, and the unchanged augmentations we left blank in Table  \ref{table:aug_fix}. Apart from modifying augmentation parameters, we also removed \textit{Contrast} and added \textit{Hue}. We kept the same augmentation parameters for weak augmentation of labeled images, and we only used \textit{horizontal flip-and-shift} and random \textit{crop} for our weak augmentation. With our tweaked version of augmentation, we got the best results on satellite images. With our augmentation parameters, the images retained more information after augmentation and, therefore, gave better results.
\vspace{-1mm}
 \begin{figure}[h]
    \centering
    \includegraphics[scale=0.52]{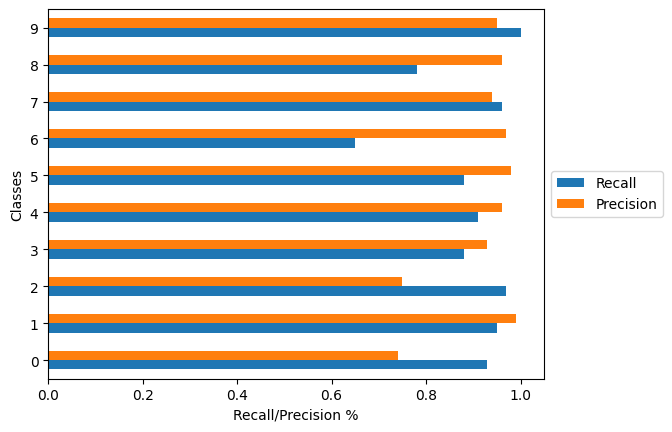}
    \caption{Precision \& Recall of a FixMatch model on EuroSAT}
    \label{fig:plots}
\end{figure}
 \subsection{Class-rebalancing}
Class-rebalancing is an iterative process; it trains the model for a couple of \textit{generations} to rebalance the imbalanced classes. Due to imbalanced classes, the model gets biased towards the majority classes and thus gives us a biased result \cite{buda2018systematic}. As shown in Figure \ref{fig:plots}, it is evident that the model is biased in favor of the majority class. Despite its bias, the model maintains its precision in minority classes. The recall is significantly lower in the minority class and higher in the majority class.

In this section, we discuss how our proposed method handles class imbalance. If the pseudo-label is confident enough and belongs to minority classes, the pseudo-labeled image gets added to the labeled set. Now, the pseudo-labels from minority classes are added at a higher sampling rate to increase the number of samples in minority classes of the labeled set.

\begin{equation}
\mu_\textit{l} = \left(\frac{N_{L+1-\textit{l}}}{N_1}\right)^\alpha\\\label{eq:1}
\end{equation}
This sampling
rate is decided by a predefined sampling hyperparameter tuner $\alpha= 1/3$. And with the
the help of equation \ref{eq:1}  we calculate the adaptive sampling rate.
\begin{figure}[h]
    \centering
    \includegraphics[width=0.40\textwidth]{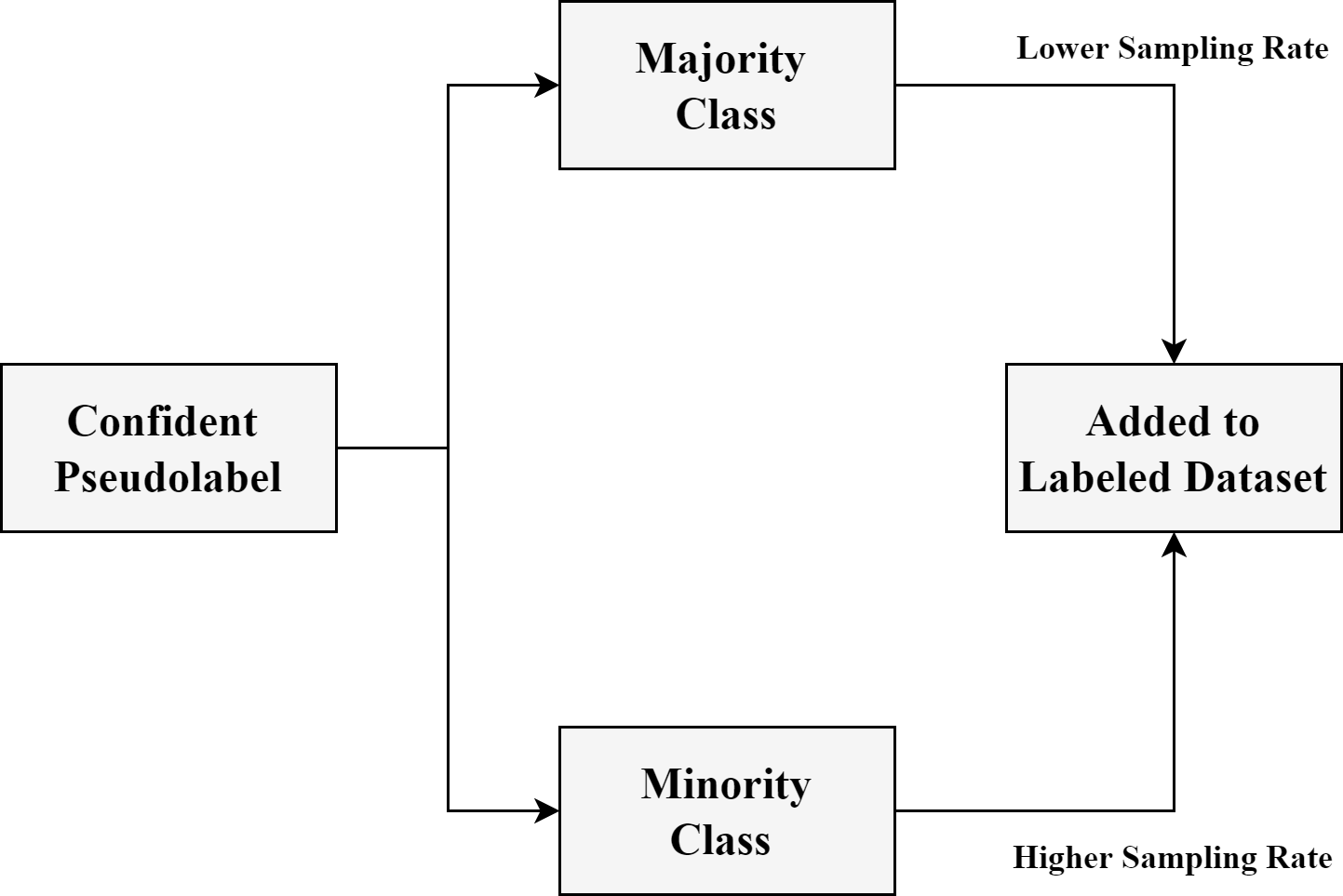}
    \caption{ Class Rebalancing pipeline for the algorithm}
    \label{fig:Distribution}
\end{figure}

From figure \ref{fig:Distribution}, we can see that the minority class pseudo-labels should be added at a higher sampling rate. This is supported by the result from equation \ref{eq:1}, where we get a higher sampling rate for minority classes and a lower sampling rate for majority classes. After getting added to the labeled set, the model gets retrained with the newly added data and the previous data and follows the same pipeline. By finishing each iteration, the algorithm completes one \textit{generation}, and new pseudo-labels are added in each \textit{generation}, and the model gets retrained. As new pseudo-labels are added to minority classes, the labeled set gets more balanced in each \textit{generation}, reducing the model’s bias.
 \vspace{-2mm}
\section{Results}\label{sec:result}
 \vspace{-2mm}
\subsection{Dataset}
We used three datasets, EuroSAT, UCM, and WHU-RS19, for our experiment. UCM and EuroSAT are two of the most widely used SSL benchmarks in remote sensing. Here, we primarily used EuroSAT since it has imbalanced classes, and we want to apply class rebalancing to it. So, EuroSAT is our prime example.

\textit{1) EuroSAT:} EuroSAT \cite{helber2019eurosat} is a dataset of satellite images covering 10 classes with a total of 27,000 images. Each image has a resolution of 64 × 64, which is in RGB and also in 13-band MS format. All are labeled and geo-referenced images. The 10 classes are Industrial Buildings, Residential Buildings, Annual Crop, Permanent Crop, River, Sea and Lake,  Herbaceous Vegetation, Highway, Pasture, and Forest. These datasets are suitable for deep learning models. We collected the data from the authors' GitHub repository\footnote{https://github.com/phelber/EuroSAT}.

\textit{2) UCM:}  The UCM dataset \cite{yang2010bag} consists of 21 classes, where each class has  100 land use images of 256 × 256 pixels, which makes 2100 images in total. This dataset is commonly used for scene classification. We collected the data from the authors' website\footnote{ http://weegee.vision.ucmerced.edu/datasets/landuse.html}.

\textit{3) WHU-RS19:} WHU-RS19 \cite{dai2010satellite} \cite{xia2010structural} is a collection of high-resolution satellite photos up to 0.5 m that were extracted from Google Earth. Airport, beach, bridge, commercial, desert, farmland, mountain, football field, viaducts, etc., are among the 19 classifications of essential scenes in high-resolution satellite images. There are around 50 to 61 samples for each class. There are a total of 1,013 images in the WHU-RS19 dataset. We collected the data from the authors' website\footnote{https://captain-whu.github.io/BED4RS/}.
\vspace{-3.4mm}
\subsection{Performance Analysis}
\vspace{-1mm}
Several methods were chosen as appropriate comparisons to the proposed method. We accomplish our comparisons in two sections. We used balanced and imbalanced datasets for comparative analysis to evaluate the proposed method properly.

In Table \ref{tab:2}, we evaluate our proposed method with fully-supervised learning method \cite{xia2017aid}\cite{helber2019eurosat}, MSMatch\cite{gomez2021msmatch} and FixMatch with tweaked augmentation.

\begin{table}[h]
\begin{threeparttable}
\caption{Performance comparison between the proposed method and other methods}
\label{tab:2}
\setlength\tabcolsep{0pt}

\begin{tabular*}{\columnwidth}{@{\extracolsep{\fill}} ll ccccccc}
\toprule
     Methods\tnote{a}& \multicolumn{2}{c}{Balanced} &  \multicolumn{3}{c}{Imbalanced} \\ \cmidrule{2-3} \cmidrule{4-6}
                                & UCM & WHU-RS19  & EuroSAT & UCM & WHU-RS19\\
\midrule
     Supervised\cite{xia2017aid} & 95.02 & 96.24 & 98.57 & - & - \\
     MS Match & 94.13 & - & 96.04 & - & - \\
     Fixmatch With TA & 94.74 & 93.48 & 96.13 & 92.65 & 91.25 \\

\addlinespace
    Proposed Method \\(1st Gen) & 94.93 & 93.50 & 96.75 & 94.95 & 93.55\\
    Proposed Method \\(2nd Gen) & 95.34 & 93.51 & 97.12 & 94.97 & 93.56\\
\bottomrule
\end{tabular*}

\smallskip
\scriptsize
\begin{tablenotes}
\item[a] TA: Tweaked Augmentation
\end{tablenotes}
\end{threeparttable}
\end{table} 

Fully-supervised learning method in \cite{xia2017aid} used CaffeNet on UCM and WHU-RS19 and achieved an accuracy of 95.02\% and 96.24\% with just 80\% and 60\% labels, respectively. In the case of imbalanced EuroSAT, the supervised learning method from \cite{helber2019eurosat} used a fine-tuned ResNet-50 convolutional neural network on EuroSAT and gained an accuracy of 98.57\%.

MSMatch \cite{gomez2021msmatch} uses a semi-supervised approach with EfficientNet achieving accuracies of 94.13\% and 96.04\% on UCM and EuroSAT. 

We also took the state-of-the-art semi-supervised method FixMatch, made custom augmentations, and applied it to UCM, WHU-RS19, and EuroSAT.

Then we used it to evaluate our proposed method. Comparing the performance accuracies, we see that our method using just two \textit{generations} performs (95.34\%) better than the fully-supervised method (95.02\%) and also outperforms MSMatch\cite{gomez2021msmatch} and FixMatch with the tweaked augmentation by 1.21\% and 0.6\% on the UCM balanced dataset. In the case of imbalanced EuroSAT, our method outperforms MSMatch and FixMatch by 1.08\% and 1\%, respectively. Moreover, it comes close to the fully-supervised method as well. For both balanced and imbalanced UCM and WHU-RS19, our method (with 2 \textit{generations}) outperforms them in accuracy. Furthermore, in the case of balanced UCM and WHU-RS19, our method comes close to a fully-supervised performance.
\vspace{-2.5mm}
\subsection{Training}
Our models have been trained on Kaggle's provided notebook environment with a Tesla P100-PCIE-16GB using PyTorch 1.9.1. For training, we used a stochastic gradient descent (SGD) optimizer \cite{sutskever2013importance} with a 0.9 Nesterov momentum and multiple weight decay rates. We experimented with Wide ResNet-28-2 \cite{zagoruyko2016wide} as the backbone of our model. The learning rate was set to 0.03. The training batch size was 16 for all the datasets we used. The training ran for 512 epochs with 1024 iterations for EuroSAT, UCM, and WHU-RS19 each. We used the datasets' mean and standard deviation to normalize each image.
For the test sets, we took 10\% of the dataset for UCM, EuroSAT, and WHU-RS19. EuroSAT was already imbalanced. However, we created an artificial imbalance for the class-balanced datasets, UCM and WHU-RS19. For a fair comparison between all these datasets, we kept the imbalance ratio to 0.1. A single run of EuroSAT with three \textit{generations} took over 86 hours, and for UCM and WHU-RS19, it took 205 and 175 hours, respectively, on the notebook environment.

\section{Conclusion}\label{sec:conclusion}
We developed a technique that uses semi-supervised learning to label remote sensing data and a class-rebalancing distribution technique to handle imbalanced data. Our proposed methodology helps resolve the problem of manually labeling data and also the problem of model bias brought on by data imbalance. We tweaked the augmentation strategy inspired by FixMatch\cite{sohn2020fixmatch}. which helped improve the accuracy of the satellite image datasets. We also included a class-rebalancing approach, which balances the dataset classes by aligning the class distribution by adding more samples to the classes with fewer samples. This lessens the impact of model bias. Comparing our proposed method against FixMatch with tweaked augmentation, MSMatch\cite{gomez2021msmatch}, and some supervised approaches, it is seen that our method performs better.

The datasets we have worked on are all lacking labeled images. Therefore, one of the limitations of our suggested method is that we will not be able to use datasets that contain no labeled images, i.e., all images are unlabeled because our method only needs a small amount of labeled data. Moreover, our method requires an enormous amount of unlabeled data, which is another drawback of our research. Therefore, using our suggested method, datasets containing a small proportion of unlabeled data will not produce good results. Even though we have worked with datasets with high-resolution images, we intend to work with those with even higher-resolution images in the future.
Additionally, we wish to cooperate with massive datasets like BigEarthNet \cite{long2021creating}, and Million-AID \cite{sumbul2020bigearthnet}. We also want to apply our proposed approach to LULC change analysis across different periods in a particular region. Since we have so far only worked with RGB satellite photos, we want to test how well our model will perform with multispectral (MS) images, which have additional bands and information.

\bibliographystyle{IEEEtran}
\bibliography{bibliography}

\end{document}